\documentclass[10pt,twocolumn,letterpaper]{article}

\usepackage{cvpr}
\usepackage{times}
\usepackage{epsfig}
\usepackage{graphicx}
\usepackage{amsmath}
\usepackage{amssymb}
\usepackage{subfigure}
\usepackage{color}
\usepackage{multirow}
\usepackage{diagbox}
\usepackage{array}
\usepackage{booktabs}
\usepackage[table]{xcolor}
\makeindex

\graphicspath{{./figs/}}
\def\eg{\emph{e.g.~}}
\def\etal{{\em et al.~}}

\usepackage{pifont}
\usepackage{overpic}
\newcommand{\cmark}{\ding{51}}%

\newcommand{\figref}[1]{Fig.~\ref{#1}}
\newcommand{\tabref}[1]{Table~\ref{#1}}

\newcommand{\secref}[1]{Sec.~\ref{#1}}

\newcommand{\myPara}[1]{\vspace{.05in}\noindent\textbf{#1.}}

\usepackage{url}

\newcommand{\addFig}[1]{}
\newcommand{\addFigs}[1]{}

\newcommand{\sorb}[1]{{\textcolor[rgb]{0.72,0.00,0.00}{\textbf{#1}}}}

\newcommand{\sobb}[1]{{\textcolor{blue}{\textbf{#1}}}}

\definecolor{mygreen}{RGB}{0,150,0}
\definecolor{myred}{RGB}{200,0,0}
\usepackage[pagebackref=true,letterpaper=true,colorlinks=true,linkcolor=myred,citecolor=mygreen,bookmarks=false]{hyperref}

\cvprfinalcopy % *** Uncomment this line for the final submission

\def\cvprPaperID{2031} % *** Enter the CVPR Paper ID here

\ifcvprfinal\fi

\begin{document}

%%%%%%%%% TITLE
\title{A Simple Pooling-Based Design for Real-Time Salient Object Detection}

\author{Jiang-Jiang Liu$^1$\thanks{Indicates equal contributions.}
  \quad  \quad Qibin Hou$^1$\footnotemark[1] \quad \quad 
   Ming-Ming Cheng$^1$ 
  \thanks{M.M. Cheng (cmm@nankai.edu.cn) is the corresponding author.} \quad \quad
   Jiashi Feng$^2$  \quad \quad  Jianmin Jiang$^3$\\
   $^1$TKLNDST, College of CS, Nankai University \quad \quad
   $^2$NUS  \quad \quad $^3$Shenzhen University\\
   {\tt\small  \{j04.liu, andrewhoux\}@gmail.com}
}

\maketitle
%\thispagestyle{empty}

%%%%%%%%% ABSTRACT
\begin{abstract}

We solve the problem of salient object 
detection by investigating
how to expand the role of pooling in convolutional neural networks.
Based on the U-shape architecture, we first build a global guidance module (GGM) upon
the bottom-up pathway, aiming at providing layers at different feature levels the 
location information of potential salient objects.
We further design a feature aggregation module (FAM) to
make the coarse-level semantic information well fused with
the fine-level features from the top-down pathway.
By adding FAMs after the fusion operations in the top-down pathway, coarse-level features from
the GGM can be seamlessly merged with features at various scales.
These two pooling-based modules allow the high-level semantic features to
be progressively refined, yielding detail enriched saliency maps.
Experiment results show that our proposed approach
can more accurately locate the salient objects with sharpened details
and hence substantially improve the performance compared to the previous state-of-the-arts.
Our approach is fast as well and can run at a speed of more than 30 FPS 
when processing a $300 \times 400$ image.
Code can be found at \url{http://mmcheng.net/poolnet/}.

\end{abstract}

%%%%%%%%% BODY TEXT
\section{Introduction} \label{sec:introduction}

Benefiting from the capability of detecting the most visually distinctive objects from a given image,
salient object detection plays an important role in many computer vision tasks, such as
%such as scene classification \cite{ren2014region}, image segmentation \cite{donoser2009saliency}, 
visual tracking \cite{hong2015online}, 
content-aware image editing \cite{cheng2010repfinder},
%image retrieval \cite{gao20123}, 
%person identification \cite{zhao2013unsupervised}, 
and robot navigation \cite{craye2016environment}.
Traditional methods \cite{itti1998model,liu2011learning,klein2011center,
perazzi2012saliency,borji2012exploiting,jiang2013salient,yan2013hierarchical,cheng2015global} mostly rely on hand-crafted features to
capture local details and global context separately or simultaneously, but the lack of
high-level semantic information restricts their ability to detect the integral salient objects
in complex scenes.
%
%These early works have guided and promoted the development of salient object detection greatly.
%
Luckily, convolutional neural networks (CNNs) greatly promote the development of
salient object detection models because of their capability of extracting both high-level semantic 
information and low-level detail features in multiple scale space.

As pointed out in many previous approaches \cite{hou2016deeply,luo2017non,zhang2017amulet},
because of the pyramid-like structural characteristics of CNNs, shallower stages usually have
larger spatial sizes and keep rich, detailed low-level information while deeper stages
contain more high-level semantic knowledge and are better 
at locating the exact places of salient objects.
Based on the aforementioned knowledge, a variety of new
architectures \cite{hou2016deeply,li2017instance,wang2018detect,hou2018three} for salient object detection have been designed.
Among these approaches, U-shape based structures \cite{ronneberger2015u,lin2017feature} 
receive the most attentions due to their ability to construct enriched feature maps 
by building top-down pathways upon classification networks.

Despite the good performance achieved by this type of approaches, 
there is still a large room for improving it.
First, in the U-shape structure, high-level semantic information is progressively transmitted
to shallower layers, and hence the location information captured by deeper layers may be gradually diluted
at the same time. %\textcolor{red}{[JS: not clear how the introduced pooling operation solves this issue.]}
Second, as pointed out in \cite{zhao2016pyramid}, the receptive field size of a CNN is
not proportional to its layer depth.
Existing methods solve the above-mentioned problems by introducing attention mechanisms \cite{zhang2018progressive,liu2018picanet} into U-shape structures, refining feature maps
in a recurrent way \cite{liu2016dhsnet,zhang2018progressive,wangsaliency},
combining multi-scale feature information \cite{hou2016deeply,luo2017non,zhang2017amulet,hou2018three},
or add extra constraints to saliency maps like the boundary loss term in \cite{luo2017non}.

\begin{figure*}[tp]
  \centering
  \begin{overpic}[width=1.0\linewidth]{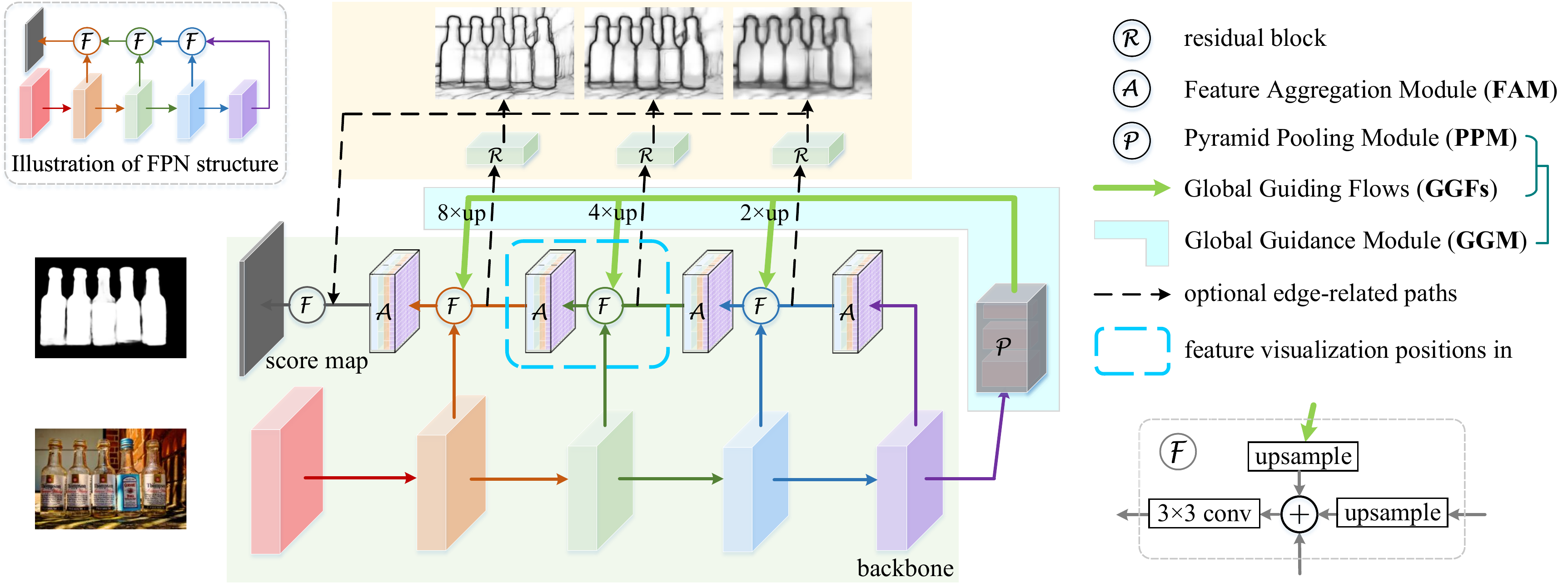}
    \put(96.9,14.65){\footnotesize \figref{fig:chm_effects}}
  \end{overpic}
  \caption{The overall pipeline of our proposed approach. For clarity, we also place a standard
  U-shape FPN structure \cite{lin2017feature} at the top-left corner. The top part for edge detection
  is optional.
  }\label{fig:arch}
\end{figure*}

% As a result, how to maintain the influence of high-level semantic information and  
% effectively enlarge the receptive field size of
% a network is essential for improving the performance of segmentation-like models.
%

In this paper, different from the methods mentioned above, 
we investigate how to solve these problems by expanding the role 
of the pooling techniques in U-shape based architectures.
In general, our model consists of two primary modules on the 
base of the feature pyramid networks (FPNs) \cite{lin2017feature}: 
a global guidance module (GGM) and a feature aggregation module (FAM).
As shown in \figref{fig:arch}, our GGM composes of a modified version 
of pyramid pooling module (PPM)
and a series of global guiding flows (GGFs).
Unlike \cite{wang2017stagewise} which directly plugs 
PPM into the U-shape networks, our GGM is an individual module.
More specifically, the PPM is placed on the top of the backbone 
to capture global guidance information (where
the salient objects are).
By introducing GGFs, high-level semantic information collected by PPM 
can be delivered to feature maps at all pyramid levels, 
remedying the drawback of U-shape networks that top-down signals
are gradually diluted.
Taking into account the fusion problem of the coarse-level feature maps from GGFs
with the feature maps at different scales of the pyramid, 
we further propose a feature aggregation module (FAM), 
which takes the feature maps after fusion as input.
This module first converts the fused feature maps into multiple feature spaces
to capture local context information at different scales
and then combines the information to weigh the compositions of
the fused input feature maps better.
%
% Thus seamlessly merge of the coarse-level feature maps from GGFs
% and the feature maps at different scales of the pyramid.
%\textcolor{red}{[JS: this description is too general]}
%

As both the above modules are based on the pooling techniques, we call our method PoolNet.
\emph{To the best of our knowledge, this is the first paper that 
aims at studying how to design various pooling-based modules to assist 
in improving the performance for salient object detection.}
As an extension of this work, we also equip our architecture with an edge detection branch
to further sharpen the details of salient objects by joint training 
our model with edge detection.
To evaluate the performance of our proposed approach, we report results on multiple
popular salient object detection benchmarks.
Without bells and whistles, our PoolNet
surpasses all previous state-of-the-art methods in a large margin.
In addition, we conduct a series of ablation experiments to let readers better understand
the impact of each component in our architecture on the performance and show how joint training
with edge detection helps enhance the details of the predicted results.

Our network can run at a speed of more than 30 FPS on a single NVIDIA Titan Xp GPU
for an input image with size $300 \times 400$.
When the edge branch is not incorporated, training only takes less than 6 hours 
on a training set of 5,000 images,
which is quite faster than most of the previous methods
% (at least 9 hours on 2,500 images)
\cite{liu2018picanet,zhang2018bi,luo2017non,zhang2017amulet,zhang2017learning,hou2016deeply}.
%\textcolor{red}{[JS: time for previous methods? and add reference to these methods]}.
%
This is mainly due to the effective utilization of pooling techniques.
PoolNet, therefore, can be viewed as a baseline to help ease future research
in salient object detection.

\section{Related Work}

%In this section we briefly introduce the related works of salient object detection.

% \subsection{Salient Object Detection}
%
Recently, benefiting from the powerful feature extraction capability of CNNs, most of the traditional
saliency detection methods based on hand-crafted features 
\cite{cheng2015global,jiang2013salient,li2013saliency,perazzi2012saliency}
have been gradually surpassed.
Li \etal \cite{li2015visual} used the multi-scale features extracted from a CNN to 
compute the saliency value for each super-pixel.
Wang \etal \cite{wang2015deep} adopted two CNNs, aiming at combining
local super-pixel estimation and global proposal searching together, to produce saliency maps.
Zhao \etal \cite{zhao2015saliency} presented a multi-context deep learning framework which
extracts both local and global context information by employing two independent CNNs.
Lee \etal \cite{lee2016deep} combined low-level heuristic features, 
such as color histogram and Gabor responses, with high-level features extracted from CNNs.
%
% And then a fully connected neural network was exploited to produce the saliency of each query region.
%
%These methods show the superior of deep learning based methods.
%
All these methods take image patches as the inputs of CNNs and hence
are time-consuming.
Moreover, they ignore the essential spatial information of the whole input image.

To overcome the above problems, more research attentions are put 
on predicting pixel-wise saliency maps, inspired by the 
fully convolutional networks \cite{long2015fully}.
Wang \etal \cite{wangsaliency} generated saliency prior maps using low-level cues and 
further exploited it to guide the prediction of saliency recurrently.
Liu \etal \cite{liu2016dhsnet} proposed a two-stage network which produces coarse saliency maps 
first and then integrates local context information to refine them recurrently and hierarchically.
Hou \etal \cite{hou2016deeply} introduced short connections into multi-scale side outputs to capture fine details.
Luo \etal \cite{luo2017non} and Zhang \etal \cite{zhang2017amulet} 
both advanced the U-shape structures and utilized multiple levels 
of context information for accurate detection of salient objects.
Zhang \etal \cite{zhang2018progressive} and Liu \etal \cite{liu2018picanet} combined 
attention mechanisms with U-shape models to guide the feature 
integration process.
Wang \etal \cite{wang2018detect} proposed a network to recurrently 
locate the salient object and then refine them with local context information.
Zhang \etal \cite{zhang2018bi} used a bi-directional structure to 
pass messages between multi-level features extracted by CNNs 
for better predicting saliency maps.
Xiao \etal \cite{xiao2018deep} adopted one network to tailor the
distracting regions first and then used another network for 
saliency detection.
%
% Though significant progress have been achieved, there is still a large room for improvement 
% by combining the nature of salient object detection with the structure of neural networks.

Our method is quiet different from the above approaches.
Instead of exploring new network architectures, we investigate
how to apply the simple pooling techniques to CNNs to simultaneously 
improve the performance and accelerate the running speed.

\section{PoolNet} \label{sec:method}

It has been pointed out in 
\cite{liu2016dhsnet,hou2016deeply,wang2017stagewise,wang2018detect} that high-level semantic
features are helpful for discovering the specific locations of salient objects.
At the meantime, low- and mid-level features are also essential for improving
the features extracted from deep layers from coarse level to fine level.
Based on the above knowledge, in this section, we propose two 
complementary modules that are capable of accurately capturing 
the exact positions of salient objects and 
meanwhile sharpening their details.
% U-shape structures consider both the aforementioned points and have been widely
% adopted by many existing works.
% %
% Based on this classic architecture and the nature of salient object detection,
%
% The proposed network incorporate the characteristic of salient object detection with the
% bottom-up top-down information flow of the U-Net architecture.
% %
% In this section, we will introduce the overall pipeline of our proposed model first.
% %
% Then two novel module GGM and FAM will be described in detail which aiming at solving the problems of global context fading
% and neighbor information confusing brought by the U-Net structure.

\subsection{Overall Pipeline}

% Figure X shows the overall pipeline of our proposed approach.
%
We build our architecture based on the feature pyramid networks (FPNs) \cite{lin2017feature}
which are a type of classic U-shape architectures designed in a bottom-up and top-down manner
as shown at the top-left corner of \figref{fig:arch}.
Because of the strong ability to combine multi-level features from 
classification networks \cite{He2016,simonyan2014very}, 
this type of architectures has been widely adopted in
many vision tasks, including salient object detection.
%
% Despite so, a key problem of U-shape structures lies in that high-level
% semantic information is progressively delivered to lower layers, which makes the
% location information captured by deeper layers gradually diluted.
% %
% To overcome this drawback, a
As shown in \figref{fig:arch}, we introduce a global guidance
module (GGM) which is built upon the top of the bottom-up pathway.
By aggregating the high-level information extracted by GGM with 
into feature maps at each feature level, our goal is to explicitly 
notice the layers at different feature levels
where salient objects are.
After the guidance information from GGM is merged with the features 
at different levels, we further introduce a feature aggregation module
(FAM) to ensure that feature maps at different scales can be 
merged seamlessly.
In what follows, we describe the structures of the above mentioned 
two modules and explain their functions in detail.

\subsection{Global Guidance Module}
\renewcommand{\addFig}[1]{\includegraphics[width=0.14\linewidth]{sgm_chm/#1}}
\renewcommand{\addFigs}[1]{\addFig{#1.jpg} & \addFig{#1.png} & 
    \addFig{#1_sal_fuse_0.png} & \addFig{#1_sal_fuse_1.png} & 
    \addFig{#1_sal_fuse_x.png} & 
    \addFig{#1_sal_fuse_2.png} & \addFig{#1_sal_fuse_3.png}}
    
\begin{figure}
	\centering
  \small
	\setlength\tabcolsep{0.2mm}
	\renewcommand\arraystretch{0.6}
	\begin{tabular}{ccccccc}
       \addFigs{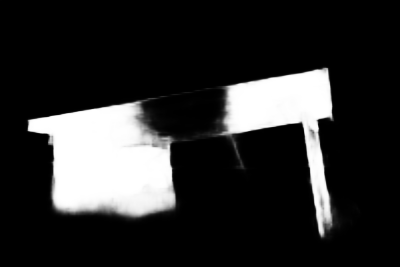}\\
		\addFigs{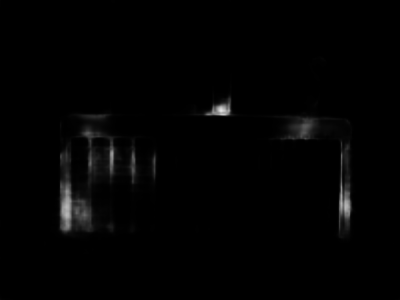}\\
		%\addFigs{sun_baxhcjlyyebupioj}\\
		\addFigs{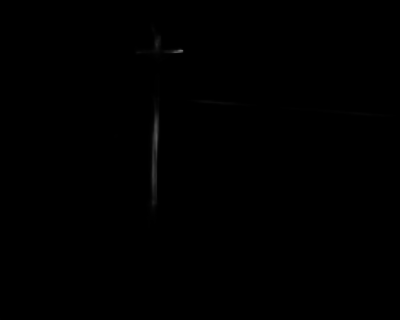}\\
		%\addFigs{sun_bikegvgcsuejydsy}\\
		%\addFigs{sun_aeyuobflqlynijyl}\\
		% \addFigs{sun_anehiuwwhtbyhaqy}\\
	   (a) & (b) & (c) & (d) & (e) & (f) & (g)\\
	\end{tabular}
  %\vspace{-8pt}
	% \includegraphics[width=\linewidth]{sqm_chm.jpg}
	\caption{Visual comparisons for salient object detection with different combinations of
    our proposed GGM and FAMs. (a) Source image; (b) Ground truth; (c) Results of FPN baseline;
    (d) Results of FPN + FAMs; (e) Results of FPN + PPM; (f) Results of FPN + GGM; (g) Results of FPN + GGM + FAMs.}
	\label{fig:module_eff_vis} 
\end{figure}
%
%In this subsection, we propose our global guidance module based on the feature pyramid networks (FPNs).
%
FPNs provide a classic architecture for combining multi-level
features from the classification backbone.
However, because the top-down pathway is built upon the bottom-up backbone, one of the
problems to this type of U-shape architectures is that the high-level features will be
gradually diluted when they are transmitted to lower layers.
%
%In addition, FPNs are composed of convolutional layers followed by non-linear transformation units.
%
It is shown in \cite{zhou2014object,zhao2016pyramid} that the empirical receptive fields 
of CNNs are much smaller than the ones in theory especially for deeper layers,
so the receptive fields of the whole networks are not large enough to capture the global
information of the input images.
The immediate effect on this is that only parts of the salient objects can be discovered as shown
in \figref{fig:module_eff_vis}c.
Regarding the lack of high-level semantic information for fine-level feature maps in the
top-down pathway, we introduce a global guidance module which contains
a modified version of pyramid pooling module (PPM) \cite{zhao2016pyramid,wang2017stagewise}
and a series of global guiding flows (GGFs) to explicitly make 
feature maps at each level be aware of the locations of 
the salient objects.

To be more specific, the PPM in our GGM consists of four sub-branches %three adaptive average pooling layers
to capture the context information of the input images.
%
%Let $F$ be the input feature map of GGM with a spatial size $H_F \times W_F$.
%
The first and last sub-branches are respectively an identity mapping layer 
and a global average pooling layer.
For the two middle sub-branches, we adopt the adaptive average pooling
layer\footnote{\url{https://pytorch.org/docs/stable/nn.html\#adaptiveavgpool2d}}
to ensure the output feature maps of them are with spatial sizes $3 \times 3$
and $5 \times 5$, respectively.
Given the PPM, what we need to do now is how to guarantee that 
the guidance information produced by PPM can be reasonably 
fused with the feature maps at different levels 
in the top-down pathway.
%
%To this end, we introduce the concept of global guiding flows (GGFs).

% As stated above, GGFs play a role of improving the capability of our network for
% localizing the most salient objects.
%
Quite different from the previous work \cite{wang2017stagewise} 
which simply views the PPM as a part of the U-shape structure, 
our GGM is independent of the U-shape structure. 
By introducing a series of global guiding flows (identity mappings),
the high-level semantic information can be easily delivered to 
feature maps at various levels (see the green arrows in
\figref{fig:arch}).
%participate in each fusion process when building the top-down pathway in FPNs.
%
%The green arrows in \figref{fig:arch} make this idea clear.
%
In this way, we explicitly increase the weight of the global guidance
information in each part of the top-down pathway to make sure 
that the location information will not be diluted when building FPNs.

To better demonstrate the effectiveness of our GGM,
we show some visual comparisons.
As depicted in \figref{fig:module_eff_vis}c, we show some saliency 
maps produced by a VGGNet version of FPNs\footnote{
Similarly to \cite{lin2017feature}, we use the feature maps outputted by {conv2, conv3, conv4, conv5}
which are denoted by $\{C_2, C_3, C_4, C_5\}$ to build the feature pyramid upon the VGGNet \cite{simonyan2014very}. The channel numbers corresponding to $\{C_2, C_3, C_4, C_5\}$
are set to $\{128, 256, 512, 512\}$, respectively.}.
It can be easily found that with only the FPN backbone, it is 
difficult to locate salient objects for some complex scenes.
There are also some results in which only parts of the salient object 
are detected.
However, when our GGM is incorporated, the quality of the 
resulting saliency maps are greatly improved.
As shown in \figref{fig:module_eff_vis}f, salient objects 
can be precisely discovered, which demonstrates the 
importance of GGM.

\subsection{Feature Aggregation Module}

\begin{figure}[tp]
  \centering
  \includegraphics[width=0.9\linewidth]{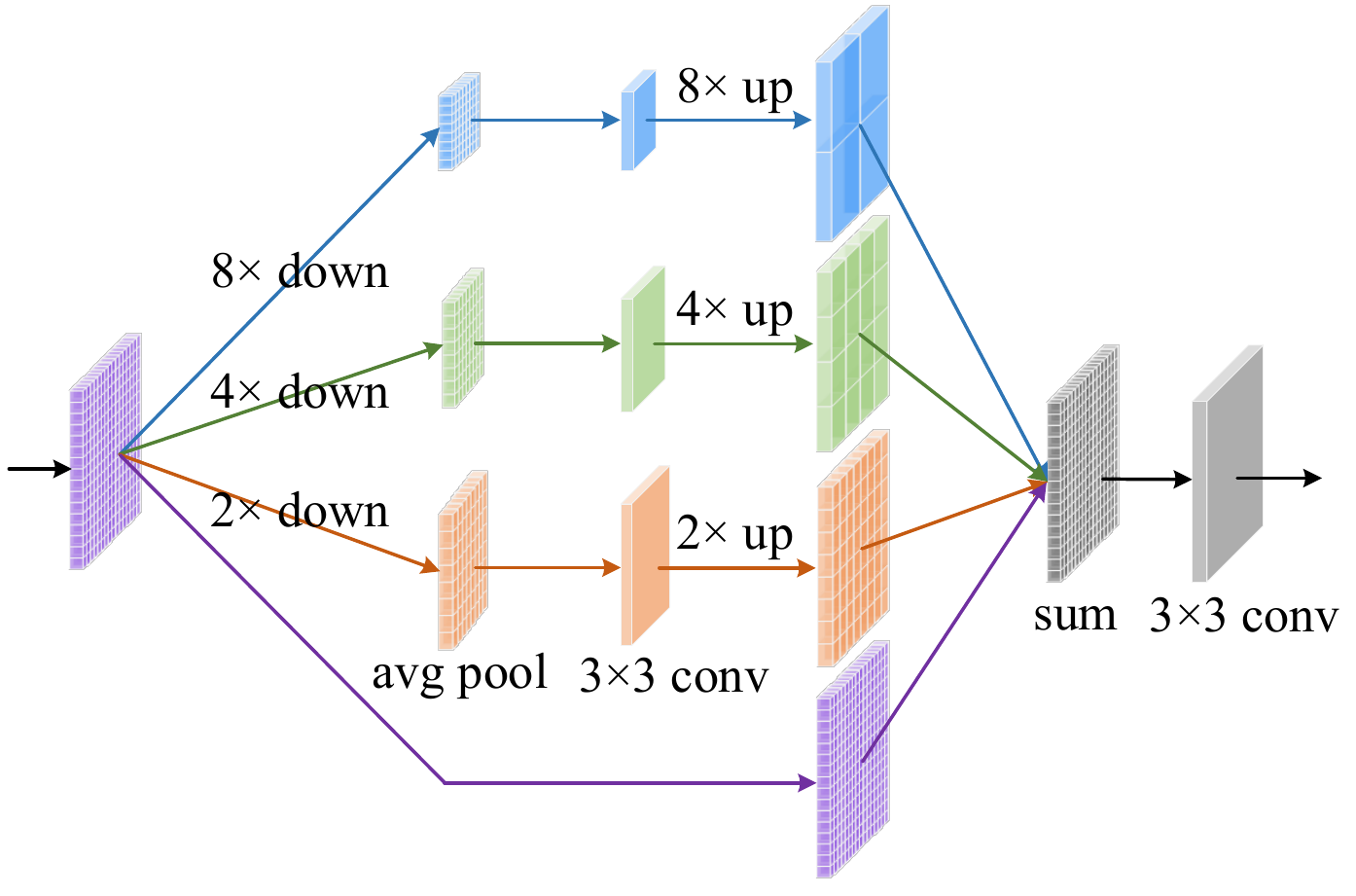}
  \caption{Detailed illustration of our feature aggregation module (FAM). It comprises
  four sub-branches, each of which works in an individual scale space. After upsampling, all
  sub-branches are combined and then fed into a convolutional layer.
  }\label{fig:fam}
  \vspace{-8pt}
\end{figure}

The utilization of our GGM allows the global guidance information 
to be delivered to feature maps at different pyramid levels.
However, a new question that deserves asking is how to make the 
coarse-level feature maps from GGM seamlessly merged with the 
feature maps at different scales of the pyramid.
Taking the VGGNet version of FPNs as an example, feature maps
corresponding to $C=\{C_2, C_3, C_4, C_5\}$ in the pyramid have
downsampling rates of \{2, 4, 8, 16\} compared to the size of 
the input image, respectively.
In the original top-down pathway of FPNs, feature maps with 
coarser resolutions are upsampled by a factor of 2.
Therefore, adding a convolutional layer with kernel size $3 \times 3$ 
after the merging operation can effectively reduce the aliasing 
effect of upsampling.
However, our GGFs need larger upsampling rates (\eg, 8).
It is essential to bridge the big gaps between GGFs and the
feature maps of different scales effectively and efficiently.
%
% When the factor is 8, we at least need a convolutional layer with 
% kernel size $8 \times 8$, which will definitely introduce a 
% large amount of learnable parameters.

To this end, we propose a series of feature aggregation modules, 
each of which contains four sub-branches as illustrated in \figref{fig:fam}.
In the forward pass, the input feature map is first converted to 
different scale spaces by feeding it into average pooling layers 
with varying downsampling rates.
The upsampled feature maps from different sub-branches are then merged together,
followed by a $3 \times 3$ convolutional layer.

\begin{figure}[tp]
  \centering
  \includegraphics[width=\linewidth]{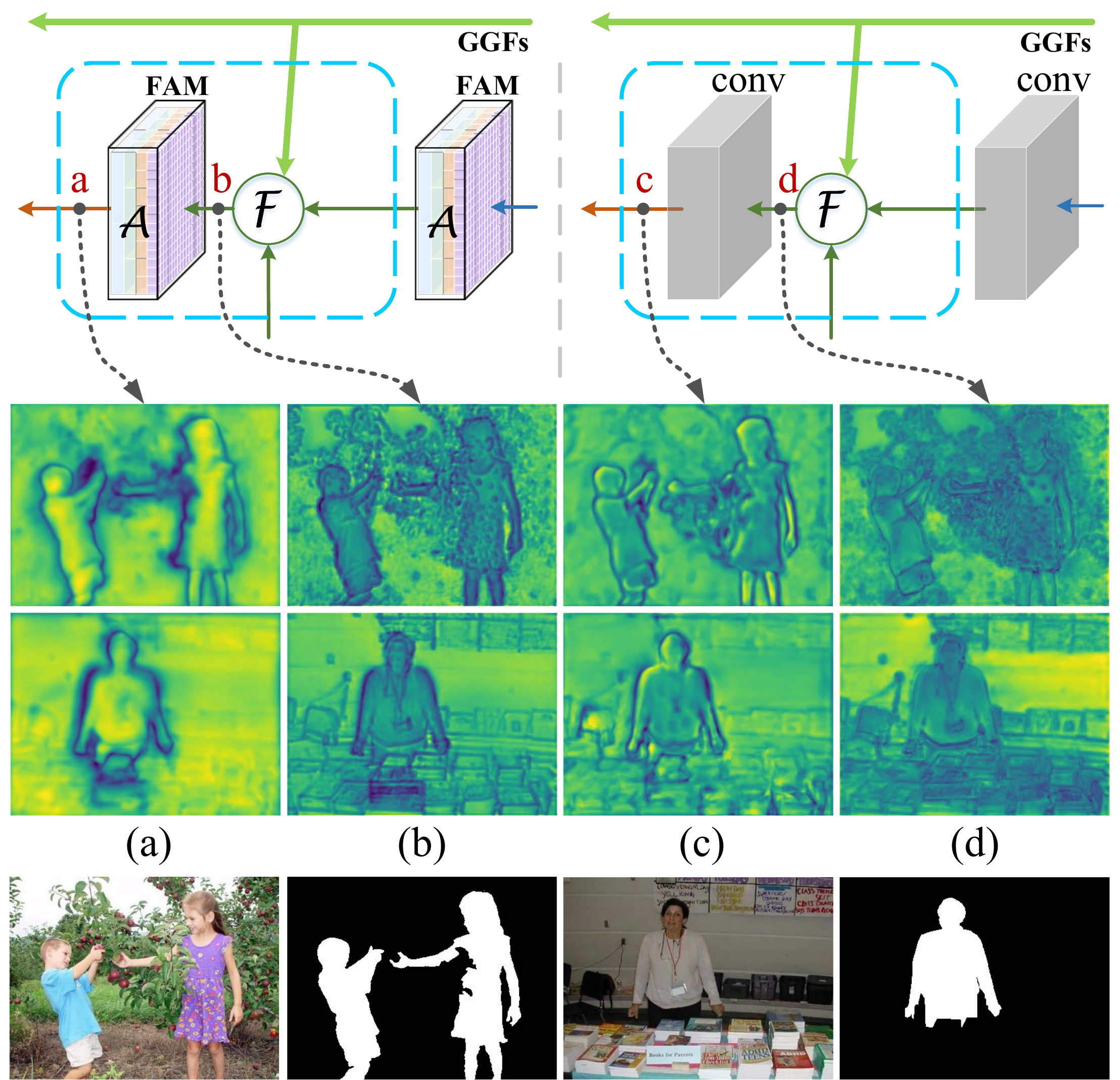} \\
  % \setlength\tabcolsep{4.8mm}
 %  \begin{tabular}{cccccc}
	%    (a) & ~~(b) & (c) & (d) & (e) & (f) \\
	% \end{tabular}
	\vspace{3pt}
	\caption{Visualizing feature maps around FAMs. Feature maps shown on the left are
	from models with FAMs, while feature maps displayed on the right
	are from models replacing FAMs with two convolution layers.
	The last row are source images and the corresponding ground-truth annotations.
	(a-d) are visualizations of feature maps at different places. As can be seen,
	when our FAMs are used, feature maps after FAMs can more precisely
	capture the location and detail information of salient objects (Column a),
	compared to those after two convolution layers (Column c).
	}
	\label{fig:chm_effects}
\end{figure}

Generally speaking, our FAM has two advantages.
%
% First, by capturing local context information at various scales,
% it balances the weights of feature maps of different scales effectively,
% assisting our model in reducing the aliasing effect of upsampling,
% especially when the upsampling rate is large (\eg, 8).
%
First, it assists our model in reducing the aliasing effect of
upsampling, especially when the upsampling rate is large (\eg, 8).
In addition, it allows each spatial location to view the local context 
at different scale spaces, further enlarging the receptive field of 
the whole network.
%
% It is worth noting that albeit our FAMs are quiet similar to the 
% structure of the PPM, \emph{quiet differently from previous works}
% \cite{zhao2016pyramid,wang2017stagewise}, we 
% our FAMs  into our model multiple times can further boost the performance.
%
\emph{To the best of our knowledge}, this is the first work revealing that FAMs
are helpful for reducing the aliasing effect of upsampling.

To verify the effectiveness of our proposed FAMs,
we visualize the feature maps near the FAMs in \figref{fig:chm_effects}.
By comparing the left part (w/ FAMs) with the right part (w/o FAMs),
feature maps after FAMs (Column a) can better capture the 
salient objects than those without FAMs (Column c).
In addition to visualizing the intermediate feature maps,
we also show some saliency maps produced by models with different settings 
in \figref{fig:module_eff_vis}.
By comparing the results in Column f (w/o FAMs) and Column g (w/ FAMs),
it can be easily found that introducing FAM multiple times allows 
our network to better sharpen the details of the salient objects.
This phenomenon is especially clear by observing the second row of \figref{fig:module_eff_vis}.
All the aforementioned discussions verify the significant effect 
of our FAMs on better fusing feature maps at different scales.
In our experiment section, we will give more numerical results.

%
% Taking the backbone of VGG-16 for example, the sizes of receptive fields of the feature
% maps generated by PPM are in the range
% of $196 \times 196$ to the full image.
% %
% And the mid- and low-level feature maps extracted from the backbone for feature maps upsampling
% are from $conv4\_3$, $conv3\_3$ and $conv2\_2$ with sizes of receptive fields of $92 \times 92$,
% $40 \times 40$ and $14 \times 14$ respectively.
% %
% This may not be of great problem in the original architecture of U-Net, which has no GGFs involved
% and the biggest stride of receptive fields difference is about 3 ($conv3\_3$ to $conv2\_2$).
% %
% But with the feature maps lead in by the GGFs, the biggest strides are now
% increasing to at least 14 ($conv5\_3$ to $conv2\_2$).
% %
% These huge gaps among receptive fields will unavoidably causing confusion in the fusion
% processes that involving both high- and low-level features.
% %
%
% We call this information confusion, which is mainly caused by inconsistent fields of view.

% %
% Simply applying a pooling operation with big kernel size and stride will still
% meet the same problem of feature conflict as a sudden blowing up of receptive
% field lacking of key information will bring no extra benefit.
%

\section{Joint Training with Edge Detection}
%
%
% \renewcommand{\addFig}[1]{\includegraphics[height=0.163\linewidth]{edge/#1}}
% % \renewcommand{\addFigs}[1]{\addFig{#1.jpg} & \addFig{#1.png} & \addFig{#1_edge_fuse_A.png} &
% % 			\addFig{#1_sal_fuse_A.png} & \addFig{#1_edge_fuse_B.png} & \addFig{#1_sal_fuse_B.png}}
% \renewcommand{\addFigs}[5]{\addFig{#1#5} & \addFig{#2#5} & \addFig{#3#5} &
% 			\addFig{#4#5}}
% \begin{figure}
% 	\centering
%     \footnotesize
% 	\setlength\tabcolsep{0.1mm}
% 	\renewcommand\arraystretch{0.6}
% 	\begin{tabular}{ccccc}
%         (a) & \addFigs{0015}{0161}{0227}{0961}{.jpg}\\
% 		(b) &\addFigs{0015}{0161}{0227}{0961}{.png}\\
% 		(c) &\addFigs{0015}{0161}{0227}{0961}{_edge_fuse_A.png}\\
% 		(d) &\addFigs{0015}{0161}{0227}{0961}{_sal_fuse_A.png}\\
% 		(e) &\addFigs{0015}{0161}{0227}{0961}{_edge_fuse_B.png}\\
% 		(f) &\addFigs{0015}{0161}{0227}{0961}{_sal_fuse_B.png}\\
% 	\end{tabular}
% 	%\vspace{-3pt}
% 	\caption{Visual examples for the impact of joint learning with edge detection. (a) Source image; (b) Ground truth; (c) salient edge map;
%     (d) salient map; (e) edge map; (f) salient map.}
% 	\label{fig:edge_impact}
% 	\vspace{-8pt}
% \end{figure}
%
%
\renewcommand{\addFig}[1]{\includegraphics[width=0.163\linewidth]{edge/#1}}
\renewcommand{\addFigs}[1]{\addFig{#1.jpg} & \addFig{#1.png} & \addFig{#1_edge_fuse_A.png} &
			\addFig{#1_sal_fuse_A.png} & \addFig{#1_edge_fuse_B.png} & \addFig{#1_sal_fuse_B.png}}
\begin{figure}
	\centering
    \footnotesize
	\setlength\tabcolsep{0.1mm}
	\renewcommand\arraystretch{0.6}
	\begin{tabular}{cccccc}
    \addFigs{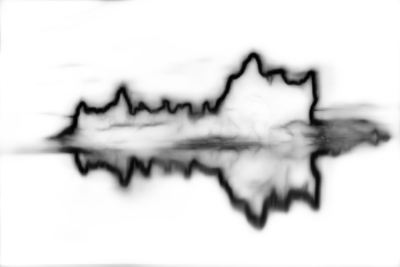}\\
		\addFigs{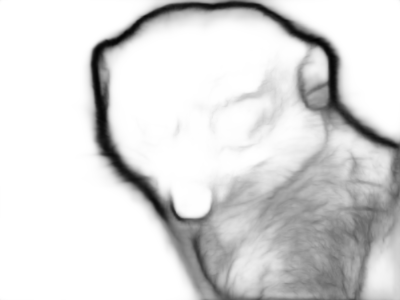}\\
		\addFigs{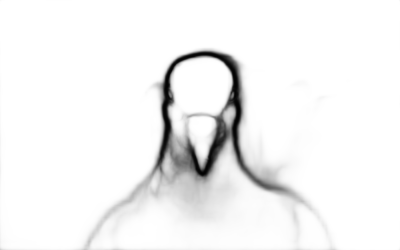}\\
		\addFigs{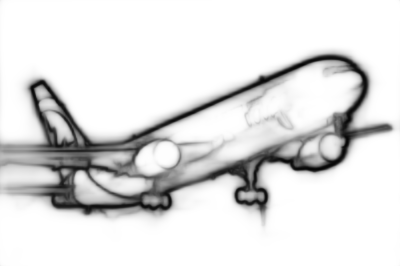}\\
		% \addFigs{0209}\\
	   (a) & (b) & (c) & (d) & (e) & (f)\\
	\end{tabular}
	\vspace{-1pt}
	\caption{Visual results by joint training with edge detection. (a) Source image; (b) Ground truth; 
	(c-d) Edge maps and saliency maps using the boundaries of salient objects as ground truths of
	the edge branch; (e-f) Edge maps and saliency maps by joint training 
	with the edge dataset \cite{arbelaez2011contour,mottaghi2014role}. By comparing the results in Column d and Column f,
	we can easily observe that joint training with high-quality edge datasets substantially improves
	the details of the detected salient objects.}
	\label{fig:edge_impact}
	\vspace{-8pt}
\end{figure}

The architecture described in \secref{sec:method} has already surpassed all previous
state-of-the-art single-model results on multiple popular salient object detection benchmarks. 
Despite so, by observing the resulting saliency maps produced by our model,
we find out that many inaccurate (incomplete or over-predicted) predictions are caused
by unclear object boundaries.

At first, we attempt to solve this problem by adding an extra prediction branch
built upon the architecture presented in \secref{sec:method} to estimate the boundaries
of the salient objects.
The detailed structure can be found on the top side of \figref{fig:arch}.
We add three residual blocks \cite{He2016} after the FAMs at three feature levels
in the top-down pathway, which are used for information transformation.
These residual blocks are similar to the design 
in \cite{He2016} and have channel numbers of 
$\{128, 256, 512\}$ from the fine level to the coarse level.
As done in \cite{liu2016richer}, each residual block is 
then followed by a 16-channel $3\times3$ convolutional layer 
for feature compression plus a one-channel $1\times1$
convolutional layer for edge prediction.
We also concatenate these three 16-channel $3\times3$
convolutional layers and feed them to three consecutive 
$3\times3$ convolutional layers with 48 channels to 
transmit the captured edge information to the salient 
object detection branch for detail enhancement. 

Similar to \cite{li2017instance}, during the training phase, 
we use the boundaries of the salient objects as our 
ground truths for joint training. 
However, this procedure does not bring us any performance gain, 
and some results are still short of detail information of 
the object boundaries.
For example, as demoed in Column c of \figref{fig:edge_impact}, 
the resulting saliency maps and boundary maps are still 
ambiguous for scenes with low contrast between the foreground and background.
The reason for this might be that the ground-truth edge maps
derived from salient objects still lack most of the detailed
information of salient objects.
They just tell us where the outermost boundaries of salient
objects are, especially for cases where there are overlaps 
between salient objects.

%
% The starting point of doing so is to add an extra supervision on the edges of the salient object,
% forcing the network to pay more attention to the places where the boundaries locate.
% %
% However, the results showing of no improvement even downgrading, (c) and (d) columns of \figref{fig:edge_impact}.
% %
% We conclude this phenomenon as that even with extra attention on the edges, 
% the provided ground truths still guide the network to locate the saliency parts 
% first, meaning that those inaccurately detected salient object's edges could not be 
% precisely detected as well, as demonstrated in \figref{fig:edge_impact}.
% %

Taking the aforementioned argument into account, we attempt to 
perform joint training with the edge detection task 
using the same edge detection dataset
\cite{arbelaez2011contour,mottaghi2014role} 
as in \cite{liu2016richer}.
During training, images from the salient object detection 
dataset and the edge detection dataset are inputted alternatively.
As can be seen in \figref{fig:edge_impact}, joint training with 
the edge detection task greatly improves the details of 
the detected salient objects.
We will provide more quantitative analysis 
in our experiment section.
%Thus, instead of using 'salient edge', we involved the same settings and data of 
%common edge detection task to learn a more general edge in the manner of joint learning.
%
% To be specific, 
% %
% This dominant leading in of edge information is accomplished by concatenating $E$ with
% $c_1$, the last feature maps of salient object detection before the final prediction. 
%
%

\section{Experimental Results}

\renewcommand{\addFig}[1]{\includegraphics[width=0.0885\linewidth]{vis_comps/#1}}
\renewcommand{\addFigs}[1]{\addFig{#1.jpg} & \addFig{#1.png} & \addFig{#1_Ours.png} &
			\addFig{#1_PiCA.png} & \addFig{#1_DGRL.png} & \addFig{#1_PAGR.png} & \addFig{#1_SRM.png} &
			\addFig{#1_AMU.png} & \addFig{#1_DSS.png} & \addFig{#1_MSR.png} & \addFig{#1_DCL.png}}
\begin{figure*}
	\centering
    \footnotesize
	\setlength\tabcolsep{0.2mm}
	\renewcommand\arraystretch{0.8}
	\begin{tabular}{ccccccccccc}
    \addFigs{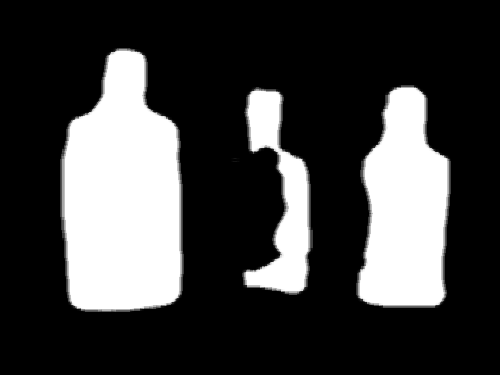}\\
		% \addFigs{166}\\
		\addFigs{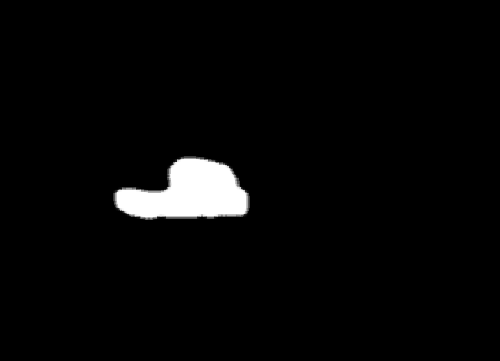}\\
		% \addFigs{0820}\\
		\addFigs{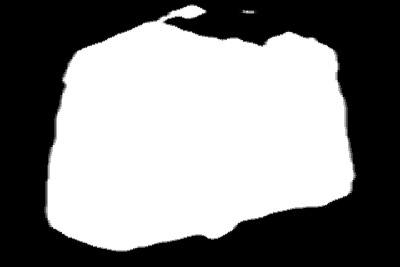}\\
		% \addFigs{0981}\\
		\addFigs{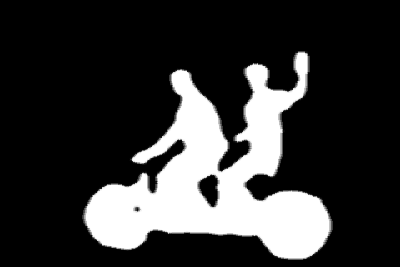}\\
		\addFigs{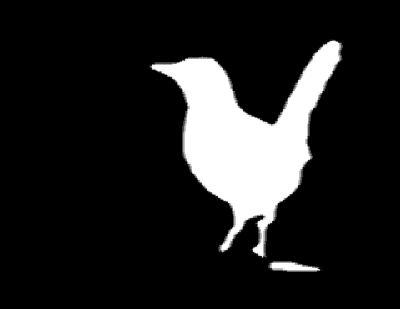}\\
	   Image & GT & Ours & PiCANet\cite{liu2018picanet} & DGRL\cite{wang2018detect} & PAGR\cite{zhang2018progressive} &
	    SRM\cite{wang2017stagewise} & Amulet\cite{zhang2017amulet} & DSS\cite{hou2016deeply} &
	    MSR\cite{li2017instance} & DCL\cite{li2016deep} \\
	\end{tabular}
  \vspace{-1pt}
	\caption{Qualitative comparisons to previous state-of-the-art methods.
	Obviously, compared to other methods, our approach 
	is capable of not only locating the integral 
	salient objects but also refining the details of the 
	detected salient objects.
	This makes our resulting saliency map very close to 
	the ground-truth annotations.}
	\label{fig:vis_comps}
  %\vspace{-8pt}
\end{figure*}

In this section, we first describe the experiment setups, including the implementation details, the used datasets and the evaluation metrics.
We then conduct a series of ablation studies to demonstrate 
the impact of each component
of our proposed approach on the performance.
At last, we report the performance of our approach and compare 
it with previous state-of-the-art methods.

\subsection{Experiment Setup}

\myPara{Implementation Details}
The proposed framework is implemented based on the PyTorch repository\footnote{\url{https://pytorch.org}}.
All the experiments are performed using the Adam \cite{kingma2014adam} optimizer with a weight decay of
5e-4 and an initial learning rate of 5e-5 
which is divided by 10 after 15 epochs.
Our network is trained for 24 epochs in total.
The backbone parameters of our network (\eg, VGG-16 \cite{simonyan2014very} and 
ResNet-50 \cite{He2016}) are initialized with the corresponding models pretrained on 
the ImageNet dataset \cite{krizhevsky2012imagenet} 
and the rest ones are randomly initialized.
By default, our ablation experiments are performed based on 
the VGG-16 backbone and the union set of MSRA-B \cite{liu2011learning} and HKU-IS \cite{li2015visual} datasets 
as done in \cite{li2017instance} unless special explanations.
We only use the simple random horizontal flipping for data augmentation.
In both training and testing, the sizes of the input images are kept unchanged as done in \cite{hou2016deeply}.
%During both training and testing, the images are fed in of their original sizes.

\myPara{Datasets \& Loss Functions}
To evaluate the performance of our proposed framework, we conduct experiments on
6 commonly used datasets, including ECSSD \cite{yan2013hierarchical}, 
PASCAL-S \cite{li2014secrets}, DUT-OMRON \cite{yang2013saliency}, 
HKU-IS \cite{li2015visual}, SOD \cite{movahedi2010design} and DUTS \cite{wang2017learning}.
Sometimes, for convenience, we use the initials of the datasets as their abbreviations
if there is no explicit conflict.
We use standard binary cross entropy loss for salient object detection and 
balanced binary cross entropy loss \cite{xie2015holistically} for edge detection.

\myPara{Evaluation Criteria}
We evaluate the performance of our approach and other methods using three
widely-used metrics: precision-recall (PR) curves, F-measure score, 
and mean absolute error (MAE).
F-measure, denoted as $F_\beta$, is an overall performance measurement and is computed 
by the weighted harmonic mean of the precision and recall:
\begin{equation}
	F_\beta = \frac{(1+\beta^2) \times Precision \times Recall}{\beta^2 \times Precision + Recall }
\end{equation}
where $\beta^2$ is set to 0.3 as done in previous work to weight precision more than recall.
The MAE score indicates how similar a saliency map $S$ is compared to the ground truth $G$:
\begin{equation}
MAE = \frac{1}{W \times H} \sum_{x=1}^W \sum_{y=1}^H \lvert S(x,y) - G(x,y) \rvert
\end{equation}
where $W$ and $H$ denote the width and height of $S$, respectively.

\renewcommand{\addFig}[1]{\includegraphics[trim = 22mm 73mm 31mm 72mm, clip, width=0.3\linewidth]{prs/#1.pdf}}
\begin{figure*}
	\centering
	\footnotesize
	\setlength\tabcolsep{1.4mm}
	\includegraphics[width=.9\linewidth]{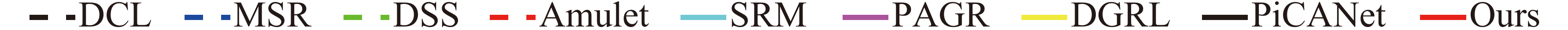}
	\begin{tabular}{cccc}
		\addFig{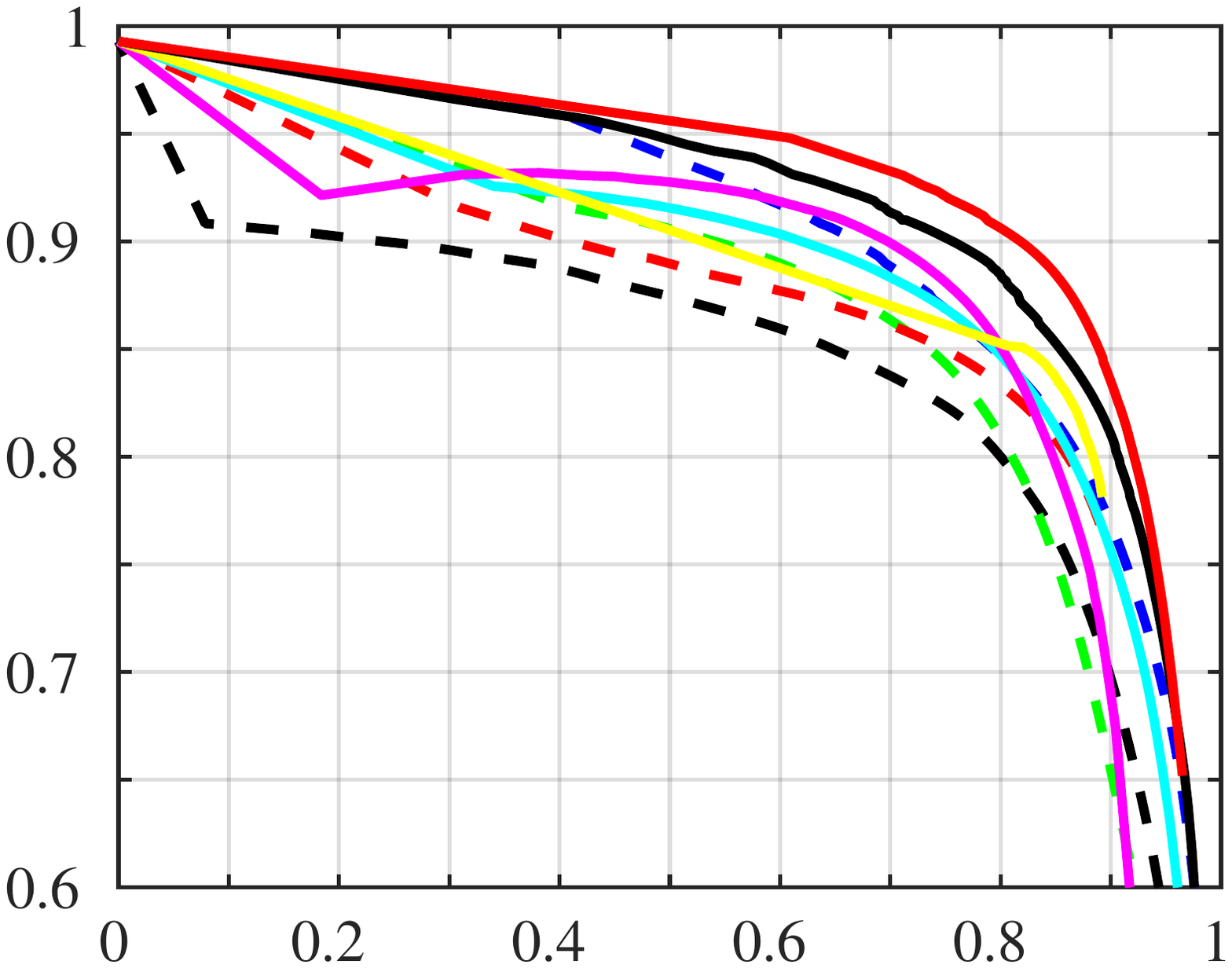} & \addFig{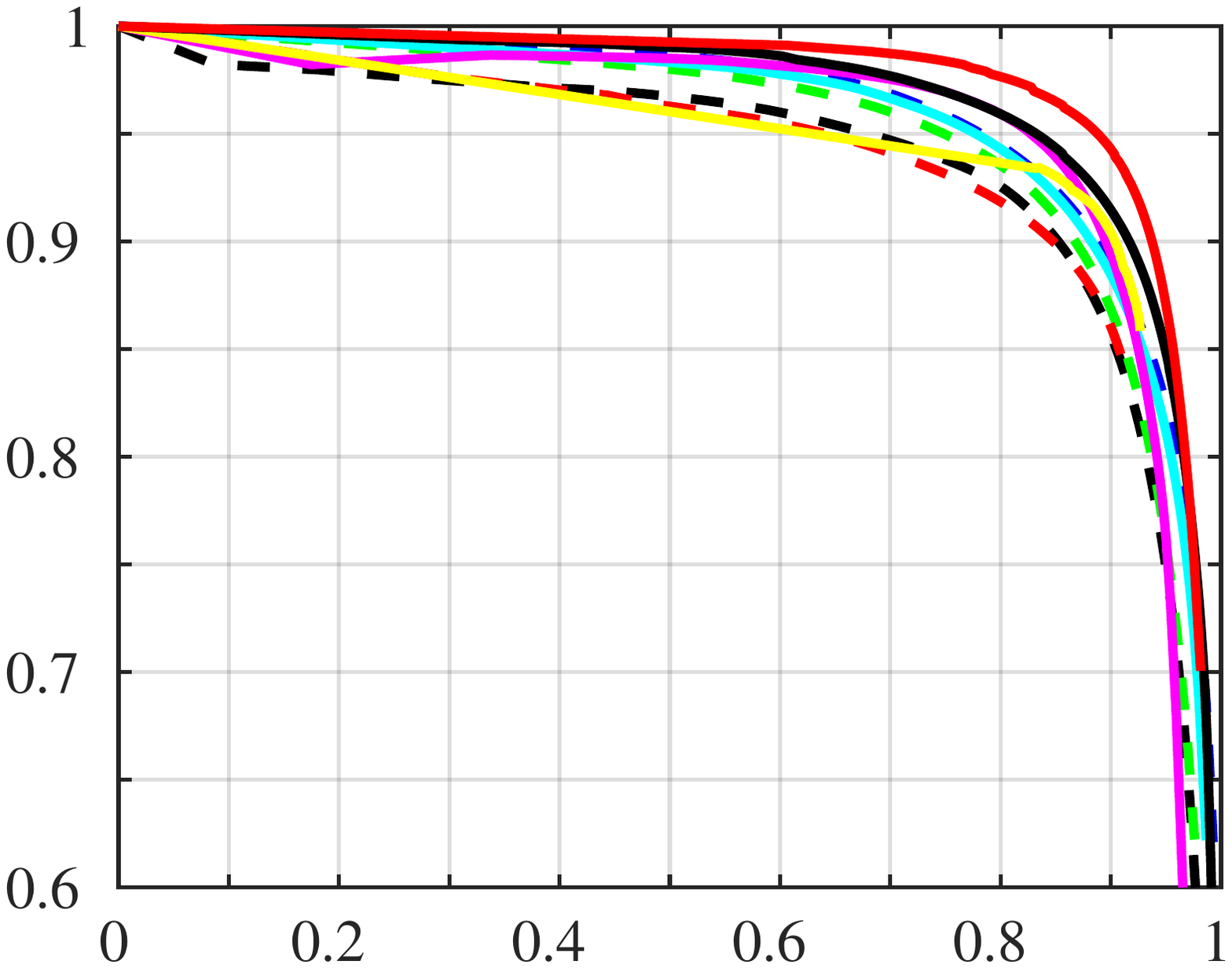} & \addFig{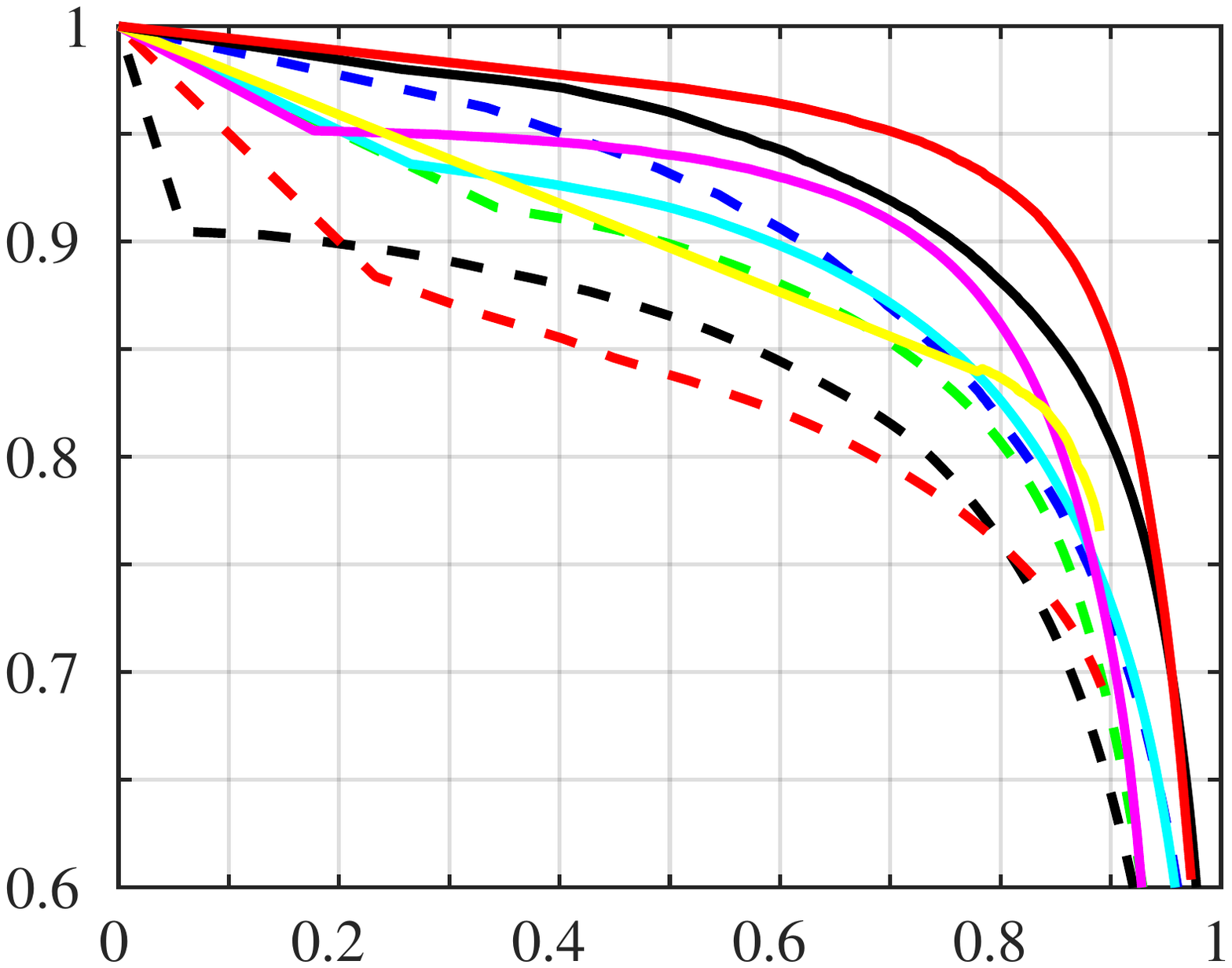} \\
		~~~~(a) PASCAL-S~\cite{li2014secrets} & ~~~~(b) HKU-IS~\cite{li2015visual} & ~~~~(c) DUTS-TE~\cite{wang2017learning} \\
	\end{tabular}
	\vspace{4pt}
	\caption{Precision (vertical axis) recall (horizontal axis) curves on
		three popular salient object datasets.}
	\label{fig:prs}
\end{figure*}

\subsection{Ablation Studies}

In this subsection, we investigate the effectiveness of our proposed GGM and FAMs first.
Then, we conduct more experiments on the configurations of our GGM and FAMs.
Finally, we show the effect of joint training with edge detection on the performance.

\myPara{Effectiveness of GGM and FAMs}
\begin{table}[t]
  \centering
  \small
  \setlength\tabcolsep{1.1mm}
  \begin{tabular}{c|ccc|cccc} \toprule[1pt]
    \multirow{2}*{No.} & \multicolumn{3}{c}{GGM + FAMs}  & \multicolumn{2}{c}{DUT-O \cite{yang2013saliency}} & \multicolumn{2}{c}{SOD \cite{movahedi2010design}} \\
	\cmidrule(l){2-4} \cmidrule(l){5-6} \cmidrule(l){7-8}
     & PPM & GGFs & FAMs & MaxF~$\uparrow$ & MAE~$\downarrow$ & MaxF~$\uparrow$ & MAE~$\downarrow$  \\ \midrule[1pt]
    % \rowcolor{gray!50}
    1  &        &        &        & 0.770 & 0.076 & 0.838 & 0.124  \\
    2  & \cmark &        &        & 0.783 & 0.071 & 0.847 & 0.125  \\
    3  &        & \cmark &        & 0.772 & 0.076 & 0.843 & 0.121 \\
    4  & \cmark & \cmark &        & 0.790 & 0.069 & 0.855 & 0.120 \\  \midrule[0.6pt]
    % \rowcolor{gray!50}
    5  &        &        & \cmark & 0.798 & 0.065 & 0.852 & 0.118 \\
    6  & \cmark & \cmark & \cmark & \sorb{0.806} & \sorb{0.063} & \sorb{0.861} & \sorb{0.117}  \\
    \bottomrule[1pt]
    %\multicolumn{15}{r}{\footnotesize The best performances in each column are highlighted in \sorb{red}.}
  \end{tabular}
  \vspace{1pt}
  \caption{Ablation analysis for the proposed architecture on two popular datasets.
  All experiments are based on the VGG-16 backbone and trained on the union set of
  MSRA-B \cite{liu2011learning} and HKU-IS \cite{li2015visual}. By default, our baseline is
  the VGG-16 version of FPN \cite{lin2017feature}. As can be observed, each component in our
  architecture plays an important role and contributes to the performance. Best result in
  each column are highlighted in \sorb{red}.}
  \label{tab:sgm_chm}
  \vspace{-8pt}
\end{table}
To demonstrate the effectiveness of our proposed GGM and FAMs, we conduct ablation
experiments based on the FPN baseline with the VGG-16 backbone.
Except for different combinations of GGM and FAMs, all other configurations are the same.
\tabref{tab:sgm_chm} shows the performance on two challenging datasets:
DUT-O and SOD.
The corresponding visual comparisons can be found in \figref{fig:module_eff_vis}.
\begin{itemize}
\item
\textbf{GGM Only.} The addition of GGM (the 4th row in \tabref{tab:sgm_chm}) gives
performance gains in terms of both F-measure and MAE on the two datasets over the FPN baseline.
% The second row of Tab. \ref{tab:sgm_chm} shows that the
% insertion of GGM brings promotions of 2.2\%, 2.3\%, 1.9\%, 1.7\% in terms
% of F-measure and 5.0\%, 9.2\%, 5.6\%, 5.1\% in terms
% of MAE on the four datasets over the FPN baseline, respectively.
%
% Corresponding visual comparisons can be found in the (c) and (e) columns of Fig. \ref{fig:module_eff_vis}.
%
% The global saliency localization information captured by PPM, is then progressively integrated with
% the mid- and low-level feature maps.
%
The global guidance information produced by GGM allows our network to focus more on
the integrity of salient objects, greatly improving the quality of
the resulting saliency maps.
%reduces the possibility of detecting only parts of the salient objects.
%adding undesired details to the parts of non-salient from a
%global view yet 'salient' in a local view (\eg the last 2 row of \figref{fig:module_eff_vis}).
%
Therefore, the details of the salient objects can be sharpened, which might be wrongly
estimated as background for models with limited receptive fields (\eg, the last row in \figref{fig:module_eff_vis}).
\item
\textbf{FAMs Only.} Simply embedding FAMs (the 5th row of \tabref{tab:sgm_chm}) into the FPN baseline 
as shown in \figref{fig:arch} also helps improve the performance on both F-measure and MAE scores
on the same two datasets.
% a simple insertion of FAM on the FPN baseline gains 2.8\%, 3.6\%, 1.7\%, 2.7\% of improvement
% over F-measure and 7.9\%, 14.5\%, 4.8\%, 8.5\% over MAE on the same four datasets.
%
This might be because the pooling operations inside FAMs also enlarge the
receptive field of the whole network compared to the baseline, and 
the FPN baseline still needs to merge feature maps from different levels,
which indicates the effectiveness of our FAMs for solving the aliasing effect of upsampling.
%even between adjacent level of feature maps it is 
%still useful to take multi-scale context information into account.
%
%
\item
\textbf{GGM \& FAMs.} By introducing both GGM and FAMs into the baseline 
(the last row of \tabref{tab:sgm_chm}),
the performance compared to the above two cases can be further enhanced on both
F-measure and MAE scores.
%
%This proves the effectiveness of our proposed GGM and FAM.
%
%With GGM only, the aliasing problem during cross-level feature fusion become more serious 
%while the precision of localizations is increased.
%
%With FAM only, the overall size of receptive fields is slightly increased and 
%local context information helps while the lack of global view still exists.
%small gaps between
%neighbor-level feature maps are smoothed while the lack of global view still exists.
%
This phenomenon demonstrates that our GGM and FAM are two complementary modules.
The utilization of them allows our approach to possess the strong capability of
accurately discovering the salient objects and refining the details
as illustrated in \figref{fig:module_eff_vis}.
More qualitative results can be found in \figref{fig:vis_comps} as well.

\end{itemize}
\myPara{Configuration of GGM}
To have a better understanding of the constitution of our proposed GGM,
we perform two ablation experiments, which correspond to the 2nd and 3rd rows of \tabref{tab:sgm_chm},
respectively.
We alternatively remove one of the PPM and GGFs while keeping the other one unchanged.
As can be seen, both operations make the performance decline compared to the 
results with both of them considered (the 4th row).
% which are removing the PPM and the GGFs in GGM separately
% (the 2nd and 3rd rows of \tabref{tab:sgm_chm} compared to the 4th row, respectively).
%
These numerical results indicate that both PPM and GGFs play an important role in our GGM.
The absence of any one of them is harmful for the performance of our approach.

% \myPara{The Configuration of FAM}
% %
% We set the biggest kernel size of the pooling layers to $8 \times 8$ considering
% the size limits of $C_5$, feature maps from the last layer of backbone.
% %
% As FAM is designed to smooth the stride gaps during the feature fusion process
% of GGFs with lower-level feature maps,
% when we keep only one of the pooling branches of FAM in sequence (the 6th-8th rows of \tabref{tab:sgm_chm}),
% the performances on four datasets vary quite a lot.
% %
% The combination of all pooling branches obtains the best performances (the 9th row of \tabref{tab:sgm_chm}),
% which proves the effectiveness of providing multi-scale context information during cross-level feature
% integration.
%

\begin{table}[tp]
  \centering
  \small
  \setlength\tabcolsep{1.3mm}
  \begin{tabular}{lcccccc}
    \toprule[1pt]
       & \multicolumn{2}{c}{PASCAL-S \cite{li2014secrets}} & \multicolumn{2}{c}{DUT-O \cite{yang2013saliency}} & \multicolumn{2}{c}{SOD \cite{movahedi2010design}} \\
	\cmidrule(l){2-3} \cmidrule(l){4-5} \cmidrule(l){6-7} 
    Settings & MaxF & MAE & MaxF & MAE & MaxF & MAE \\ \midrule[1pt]
    Baseline (B)  & 0.838 & 0.093 & 0.806 & 0.063 & 0.861 & 0.117  \\ %\midrule[0.5pt]
	% \textbf{B + Edge}     & 0.842 & 0.081 & 0.805 & 0.059 & 0.866 & 0.109 \\
	\textbf{B + SalEdge}     & 0.835 & 0.096 & 0.805 & 0.063 & 0.863 & 0.120 \\
    \textbf{B + StdEdge} & \sorb{0.849} & \sorb{0.077} & \sorb{0.808} & \sorb{0.059} & \sorb{0.872} & \sorb{0.105} \\
    \bottomrule[1pt]
  \end{tabular}
  \vspace{3pt}
  \caption{Ablation analysis of our approach when different kinds of boundaries are used. 
  The baseline here refers to the VGG-16 version of FPN plus GGM + FAMs.
  We also use the combination of MSRA-B \cite{liu2011learning} and HKU-IS \cite{li2015visual} as 
  the training set. `SalEdge' refers to the boundaries of salient objects and `StdEdge' refers to
  the standard datasets for edge detection, which include BSDS500 \cite{arbelaez2011contour} and PASCAL VOC Context \cite{mottaghi2014role} as done in \cite{liu2016richer,kokkinos2015pushing}.}
  \label{tab:joint_edge}
  \vspace{-8pt}
\end{table}

\begin{table*}[tp!]
  \centering
  % \small
  \footnotesize
  \renewcommand{\arraystretch}{1.1}
  \renewcommand{\tabcolsep}{1.0mm}
  \begin{tabular}{lcccccccccccccc}
  \toprule[1pt]
   & \multicolumn{2}{c}{Training} & \multicolumn{2}{c}{ECSSD \cite{yan2013hierarchical}} & \multicolumn{2}{c}{PASCAL-S \cite{li2014secrets}} & \multicolumn{2}{c}{DUT-O \cite{yang2013saliency}} & \multicolumn{2}{c}{HKU-IS \cite{li2015visual}} & \multicolumn{2}{c}{SOD \cite{movahedi2010design}} & \multicolumn{2}{c}{DUTS-TE \cite{wang2017learning}} \\
   \cmidrule(l){2-3} \cmidrule(l){4-5} \cmidrule(l){6-7} \cmidrule(l){8-9} \cmidrule(l){10-11} \cmidrule(l){12-13} \cmidrule(l){14-15}
   Model & \#Images & Dataset & MaxF~$\uparrow$ & MAE~$\downarrow$ & MaxF~$\uparrow$ & MAE~$\downarrow$ & MaxF~$\uparrow$ & MAE~$\downarrow$ & MaxF~$\uparrow$ & MAE~$\downarrow$ & MaxF~$\uparrow$ & MAE~$\downarrow$ & MaxF~$\uparrow$ & MAE~$\downarrow$ \\
  \midrule[1pt]
  \multicolumn{15}{l}{\textbf{VGG-16 backbone}} \\ \midrule[1pt]
  \textbf{DCL}~\cite{li2016deep} & 2,500 & MB & 0.896 & 0.080 & 0.805 & 0.115 & 0.733 & 0.094 & 0.893 & 0.063 & 0.831 & 0.131 & 0.786 & 0.081 \\
  \textbf{RFCN}~\cite{wangsaliency} & 10,000 & MK & 0.898 & 0.097 & 0.827 & 0.118 & 0.747 & 0.094 & 0.895 & 0.079 & 0.805 & 0.161 & 0.786 & 0.090 \\
  \textbf{DHS}~\cite{liu2016dhsnet} & 9,500 & MK+DTO & 0.905 & 0.062& 0.825 & 0.092 & - & - & 0.892 & 0.052 & 0.823 & 0.128 & 0.815 & 0.065 \\
%   \textbf{ELD}~\cite{lee2016deep} & 9,000 & MK & 0.865 & 0.082 & 0.772 & 0.122 & 0.738 & 0.093 & 0.843 & 0.072 & 0.762 & 0.154 & 0.747 & 0.092 \\
  \textbf{MSR}~\cite{li2017instance} & 5,000 & MB + H & 0.903 & 0.059 & 0.839 & 0.083 & 0.790 & 0.073 & 0.907 & 0.043 & 0.841 & 0.111 & 0.824 & 0.062 \\
  \textbf{DSS}~\cite{hou2016deeply} & 2,500 & MB & 0.906 & 0.064 & 0.821 & 0.101 & 0.760 & 0.074 & 0.900 & 0.050 & 0.834 & 0.125 & 0.813 & 0.065 \\
  \textbf{NLDF}~\cite{luo2017non} & 3,000 & MB & 0.903 & 0.065 & 0.822 & 0.098 & 0.753 & 0.079 & 0.902 & 0.048 & 0.837 & 0.123 & 0.816 & 0.065 \\
  \textbf{UCF}~\cite{zhang2017learning} & 10,000 & MK & 0.908 & 0.080 & 0.820 & 0.127 & 0.735 & 0.131 & 0.888 & 0.073 & 0.798 & 0.164 & 0.771 & 0.116 \\
  \textbf{Amulet}~\cite{zhang2017amulet} & 10,000 & MK & 0.911 & 0.062 & 0.826 & 0.092 & 0.737 & 0.083 & 0.889 & 0.052 & 0.799 & 0.146 & 0.773 & 0.075 \\
  \textbf{GearNet}\cite{hou2018three} & 5,000 & MB + H & 0.923 & 0.055 & - & - & 0.790 & 0.068 & 0.934 & 0.034 & 0.853 & 0.117 & - & - \\
  \textbf{PAGR}~\cite{zhang2018progressive} & 10,553 & DTS & 0.924 & 0.064 & 0.847 & 0.089 & 0.771 & 0.071 & 0.919 & 0.047 & - & - & 0.854 & 0.055 \\
  \textbf{PiCANet}~\cite{liu2018picanet} & 10,553 & DTS & 0.930 & 0.049 & 0.858 & 0.078 & 0.815 & 0.067 & 0.921 & 0.042 & 0.863 & \sobb{0.102} & 0.855 & 0.053 \\ \hline
  % \textbf{Ours} & 2,500 & MB & 0.921 & 0.050 & 0.846 & 0.076 & 0.790 & 0.063 & 0.913 & 0.037 & 0.853 & 0.109 & 0.832 & 0.055 \\
  \textbf{PoolNet}~\textbf{(Ours)} & 2,500 & MB & 0.918 & 0.057 & 0.828 & 0.098 & 0.783 & 0.065 & 0.908 & 0.044 & 0.846 & 0.124 & 0.819 & 0.062 \\
  % \textbf{Ours} & 5,000 & MB + H & 0.933 & 0.045 & 0.851 & 0.076 & 0.808 & 0.058 & \sobb{0.937} & \sobb{0.027} & \sobb{0.870} & 0.108 & 0.856 & 0.047 \\
  \textbf{PoolNet}~\textbf{(Ours)} & 5,000 & MB + H & 0.930 & 0.053 & 0.838 & 0.093 & 0.806 & 0.063 & \sobb{0.936} & \sobb{0.032} & 0.861 & 0.118 & 0.855 & 0.053 \\
  % \textbf{Ours} & 10,553 & DTS & \sobb{0.938} & \sobb{0.044} & 0.856 & 0.079 & 0.812 & 0.060 & 0.930 & 0.034 & \sobb{0.870} & 0.108 & 0.873 & 0.044 \\ %run-19
  \textbf{PoolNet}~\textbf{(Ours)} & 10,553 & DTS & 0.936 & 0.047 & 0.857 & 0.078 & 0.817 & 0.058 & 0.928 & 0.035 & 0.859 & 0.115 & 0.876 & 0.043 \\ %run-26
%   $\textbf{Ours}^\text{\dag}$ & 10,553 & DTS & \sobb{0.938} & \sobb{0.044} & \sobb{0.871} & \sobb{0.068} & \sobb{0.821} & \sobb{0.056} & 0.931 & 0.034 & 0.869 & 0.103 & \sobb{0.881} & \sobb{0.041} \\
  $\textbf{PoolNet}^\text{\dag}~\textbf{(Ours)}$ & 10,553 & DTS & \sobb{0.937} & \sobb{0.044} & \sobb{0.865} & \sobb{0.072} & \sobb{0.821} & \sobb{0.056} & 0.931 & 0.033 & \sobb{0.866} & 0.105 & \sobb{0.880} & \sobb{0.041} \\
  \midrule[0.8pt]
  \multicolumn{15}{l}{\textbf{ResNet-50 backbone}} \\ \midrule[1pt]
  \textbf{SRM}~\cite{wang2017stagewise} & 10,553 & DTS & 0.916 & 0.056 & 0.838 & 0.084 & 0.769 & 0.069 & 0.906 & 0.046 & 0.840 & 0.126 & 0.826 & 0.058 \\
  \textbf{DGRL}~\cite{wang2018detect} & 10,553 & DTS & 0.921 & 0.043 & 0.844 & 0.072 & 0.774 & 0.062 & 0.910 & 0.036 & 0.843 & 0.103 & 0.828 & 0.049 \\
  \textbf{PiCANet}~\cite{liu2018picanet} & 10,553 & DTS & 0.932 & 0.048 & 0.864 & 0.075 & 0.820 & 0.064 & 0.920 & 0.044 & 0.861 & 0.103 & 0.863 & 0.050 \\ \hline

  \textbf{PoolNet}~\textbf{(Ours)} & 10,553 & DTS & 0.940 & 0.042 & 0.863 & 0.075 & 0.830 & 0.055 & 0.934 & 0.032 & 0.867 & \sorb{0.100} & 0.886 & 0.040 \\
%   $\textbf{Ours}^\text{\dag}$ & 10,553 & DTS & \sorb{0.945} & \sorb{0.038} & \sorb{0.880} & \sorb{0.065} & \sorb{0.830} & \sorb{0.054} & \sorb{0.936} & \sorb{0.030} & \sorb{0.874} & 0.104 & \sorb{0.894} & \sorb{0.036} \\
  $\textbf{PoolNet}^\text{\dag}~\textbf{(Ours)}$ & 10,553 & DTS & \sorb{0.945} & \sorb{0.038} & \sorb{0.880} & \sorb{0.065} & \sorb{0.833} & \sorb{0.053} & \sorb{0.935} & \sorb{0.030} & \sorb{0.882} & 0.102 & \sorb{0.892} & \sorb{0.036} \\
  \bottomrule[1pt]
  \multicolumn{15}{r}{\scriptsize MB: MSRA-B \cite{liu2011learning}, MK: MSRA10K \cite{cheng2015global},
  DTO: DUT-OMRON \cite{yang2013saliency}, H: HKU-IS \cite{li2015visual}, DTS: DUTS-TR \cite{wang2017learning}.}
  \vspace{1pt}
  \end{tabular}
  \caption{Quantitative salient object detection results on 6 widely used datasets.
  The best results with different backbones are highlighted in \sobb{blue} and \sorb{red}, respectively.
  $^\text{\dag}$: joint training with edge detection.
  As can be seen, our approach achieves the best results on nearly all datasets in terms of F-measure and MAE.
  }\label{tab:results}
  \vspace{-8pt}
\end{table*}

\begin{table}[h]
  \centering
  \footnotesize
  \setlength\tabcolsep{0.9mm}
  \begin{tabular}{c|ccccc} \midrule[1pt]
          & Ours    & PiCANet \cite{liu2018picanet} & DGRL \cite{wang2018detect} & SRM \cite{wang2017stagewise} & Amulet \cite{zhang2017amulet} \\\hline
    Size  & $400\times300$ & $224\times224$ & $384\times384$ & $353\times353$ & $256\times256$  \\\hline
    FPS & 32      &    7    &   8    &   14    &    16    \\
    \midrule[0.7pt]
          & UCF \cite{zhang2017learning} & NLDF \cite{luo2017non}  & DSS \cite{hou2016deeply} & MSR \cite{li2017instance} & DHS \cite{liu2016dhsnet} \\\hline
    Size  & $224\times224$ & $400\times300$ & $400\times300$ & $400\times300$ & $224\times224$ \\\hline
    FPS &   23    &   12    &   12    &  2      &  23     \\
    \midrule[1pt]
  \end{tabular}
  \caption{Average speed (FPS) comparisons between our approach (ResNet-50, w/ edge) and the previous state-of-the-art methods.}
  \label{tab:time}
  \vspace{-10pt}
\end{table}

\myPara{The Impact of Joint Training}
To further improve the quality of saliency maps produced by our approach,
we attempt to combine edge detection with salient object detection in a 
joint training manner.
In \tabref{tab:joint_edge}, we list the results when two kinds of boundary information
are considered.
As can be seen, using the boundaries of salient objects as
supervision results in no improvement
while using standard boundaries can greatly 
boost the performance on all three datasets 
especially on the MAE metric.
This indicates that involving detailed edge information 
is helpful for salient object detection.

\subsection{Comparisons to the State-of-the-Arts}

In this section, we compare our proposed PoolNet with 13 previous state-of-the-art methods,
including DCL~\cite{li2016deep}, RFCN~\cite{wangsaliency}, DHS~\cite{liu2016dhsnet}, MSR~\cite{li2017instance}, DSS~\cite{hou2016deeply}, NLDF~\cite{luo2017non}, UCF~\cite{zhang2017learning},
Amulet~\cite{zhang2017amulet}, GearNet\cite{hou2018three}, PAGR~\cite{zhang2018progressive}, PiCANet~\cite{liu2018picanet},
SRM~\cite{wang2017stagewise}, and DGRL~\cite{wang2018detect}.
For fair comparisons,
the saliency maps of these methods are generated by the original code released by the authors or directly provided by them.
Moreover, all results are directly from single-model test 
without relying on any post-processing tools and all the predicted
saliency maps are evaluated with the same evaluation code.

\myPara{Quantitative Comparisons}
Quantitative results are listed in \tabref{tab:results}.
We consider both VGG-16 \cite{simonyan2014very} and 
ResNet-50 \cite{He2016} as 
our backbones and show results on both of them.
Additionally, we also conduct experiments on different training 
sets to eliminate the potential performance fluctuation.
From \tabref{tab:results}, we can observe that our PoolNet 
surpasses almost all previous state-of-the-art results 
on all datasets with the same backbone and training set.
Average speed (FPS) comparisons among different methods 
(tested in the same environment) are also reported in \tabref{tab:time}.
Obviously, our approach runs in real time and faster
than other methods.

\myPara{PR Curves}
Other than numerical results, we also show the PR curves
on three datasets as shown in \figref{fig:prs}.
As can be seen, the PR curves by our approach (red ones) 
are especially outstanding compared to all other previous 
approaches.
As the recall score approaches 1, our precision 
score is much higher than other methods.
This phenomenon reveals that the false positives 
in our saliency map are low.

\myPara{Visual Comparisons}
To further explain the advantages of our approach, we show
some qualitative results in \figref{fig:vis_comps}.
From top to bottom, the images correspond to scenes with 
transparent objects, small objects, large objects, complex texture, and low contrast between foreground and background, respectively.
% We select multiple representative images incorporating various difficult circumstances,
% including salient objects of transparency, salient objects with different sizes, complex scenes and
% low contrast between foreground and background.
%
% Surprisingly, our approach can work well for all cases 
% listed above compared to the other approaches.
It can be easily seen that our approach can not only highlight
the right salient objects but also
maintain their sharp boundaries in almost all circumstances.
% %
% More visual comparison results can be found in the supplementary material.
%

\section{Conclusion}

In this paper, we explore the potentials of pooling on salient object detection by
designing two simple pooling-based modules: global guidance module (GGM) and feature aggregation module (FAM).
By plugging them into the FPN architecture, we show that our proposed PoolNet can surpass
all previous state-of-the-art approaches on six widely-used salient object detection
benchmarks.
Furthermore, we also reveal that joint training our network with
the standard edge detection task in an end-to-end learning manner can
greatly enhance the details of the detected salient objects.
Our modules are independent of network architectures and hence can be flexibly
applied to any pyramid-based models.
These directions also provide promising ways to improve the quality of
saliency maps. 

\paragraph{Acknowledgements.}
This research was supported by NSFC (61620106008, 61572264), 
the national youth talent support program,
Tianjin Natural Science Foundation (17JCJQJC43700, 18ZXZNGX00110)
and the Fundamental Research Funds for the Central Universities 
(Nankai University, NO. 63191501).

%-------------------------------------------------------------------
{\small
\bibliographystyle{ieee}
\bibliography{sal}
}

\end{document}